\documentclass[10pt,twocolumn]{article}
\usepackage[a4paper,margin=19mm,columnsep=6mm]{geometry}
\usepackage{graphicx,amsmath,booktabs,hyperref,float}
\hypersetup{colorlinks=true,urlcolor=blue}
\newif\ifanonymous
\ifdefined\ANON\anonymoustrue\else\anonymousfalse\fi
\newsavebox{\paperTable}
\newcommand{\fitTable}[1]{\sbox{\paperTable}{#1}\ifdim\wd\paperTable>\textwidth\resizebox{\textwidth}{!}{\usebox{\paperTable}}\else\usebox{\paperTable}\fi}

\begin{document}
\title{Readout and Operating-Point Sensitivity in Event-Stream Candidate Selection}
\ifanonymous\author{Anonymous author}\else\author{Yifan Shuai\\LiAuto\\\texttt{sddsyf@gmail.com}}\fi
\date{}
\maketitle
\begin{abstract}
Event-triggered recording needs a useful combination of target capture and false triggers, yet representation comparisons can change with readout and threshold calibration. We evaluate real event accumulations from 520 DailyDVS-200 clips and 47 participants using participant-disjoint training, validation, and testing. The original study varies frozen encoder scale, accumulation duration, and temporal view count; a post-review extension adds two supervised-pretrained convolutional encoders, a 61,586-parameter trained CNN with three seeds, ten repeated participant partitions, and calibration sensitivity. At 50 ms/eight views under the primary handshake-or-fall policy, DINO-S CLS captures 101/132 targets with 40/388 false triggers, versus 78/132 and 35/388 for patch mean. DINO-B CLS has lower selected-threshold capture but higher fold-wise low-FPR partial AUC (0.746 versus 0.732). MobileNetV3-Small captures 97/132 with 53/388 false triggers, demonstrating why lightweight baselines and actual operating points matter. Across ten dependent repartitions, mean normalized partial AUC is 0.738 for DINO-B CLS and 0.674 for MobileNet. The original LeJEPA S-to-B capture difference is 8.3 points, but its multiplicity-adjusted interval includes zero. Duration and view-count effects remain policy dependent. We find no established JEPA-specific advantage. The evidence concerns offline action-defined candidate selection on previously used development data; it does not establish personal recording value, continuous false alarms per hour, or eyewear energy.\par
\end{abstract}
\par
\section{Introduction}
A personal recording system may combine audio, inertial sensing, explicit user actions, and vision to decide when to collect richer evidence. For its visual branch, missing a candidate can lose an opportunity to record; excessive triggers increase downstream activity. A high capture rate alone therefore does not establish a good trigger. The same concern applies when comparing models with independently calibrated thresholds.\par
Event cameras report asynchronous brightness changes [1]. OpenGlass and Helios motivate low-power wearable event sensing [2,3], but their device results do not measure the pipeline studied here. We investigate an earlier, offline question: how do representation choices and available event evidence affect binary candidate selection under a common grouped protocol?\par
The initial study focused on encoder size and accumulation budgets. Its readout controls revealed a larger descriptive change for one DINO configuration than the within-family size change. That observation motivates the present post-review analysis, without converting it into a hypothesis tested on fresh data. We show that representation, readout, and operating-point calibration interact: a ranking at a selected threshold need not match a ranking over the low-FPR ROC region. Lightweight supervised-pretrained models also provide a competitive alternative to larger frozen Transformers.\par
The contribution is an empirical assessment with three parts: (1) a common real-event protocol comparing readouts, model families, lightweight baselines, and accumulation budgets; (2) joint reporting of validation-calibrated capture/false triggers and fold-wise low-FPR discrimination; and (3) repeated participant grouping, threshold recalibration, and paired uncertainty analyses that make the limits of small-data conclusions inspectable. We do not propose a new encoder, event representation, training objective, or wearable device.\par
\section{Related Work}
Event-based vision includes native asynchronous processing and image-like event representations [1]. DailyDVS-200 supplies real event action recordings [4]. SMV-EAR explores multi-view spatiotemporal representations for event action recognition [5]; our independent polarity accumulations are not its alternative-axis projections, and our binary policies are not the complete multiclass benchmark.\par
I-JEPA and V-JEPA learn visual representations through embedding prediction [6,7]. LeJEPA combines cross-view invariance and isotropic Gaussian regularization [8]; DINO uses self-distillation [9]. Our four main image checkpoints are pinned third-party Open-Knowledge-AI ImageNet-1K releases [10--13]. Shared backbone configurations control some architectural factors, but training and export differences preclude objective-only causal comparisons. An official distilled V-JEPA 2.1-B image-path reference [14] is evaluated separately from temporal video inference. Supervised-pretrained MobileNetV3-Small [15] and ResNet-18 [16] broaden the baseline beyond self-supervised ViTs.\par
Wearable event systems demonstrate a broader sensing-and-inference motivation [2,3]. Our novelty claim is confined to the empirical protocol and sensitivity findings. We have neither synchronized wearer modalities nor personal-interest labels to demonstrate multimodal complementarity or subjective memory value.\par
\section{Protocol}
\subsection{Task, data, and inputs}
The selected DailyDVS-200 subset contains 520 AEDAT4 recordings from 47 participants: drinking, taking from a bag, phone use, putting into a bag, and handshake contribute 70 clips each; picking up and falling contribute 62 each; door opening contributes 46. P1 targets handshake or fall (132 targets/388 non-targets). P2 additionally targets door opening (178/342). A non-target means outside this fixed policy, not uninteresting to the wearer. Action labels supply deterministic binary supervision; language-model judgments are not labels or predictions.\par
Native event resolution is 320$\times$240. An event is \(e_i=(x_i,y_i,t_i,p_i)\). We count each polarity in the half-open interval \([u-W,u)\), render positive/negative counts as red/blue, and set green to zero:\par
\[
\begin{aligned}R=\operatorname{round}\{255[1-\exp(-h^+/3)]\},\\ B=\operatorname{round}\{255[1-\exp(-h^-/3)]\},\\ G=0.\end{aligned}
\]
The whole field of view is bilinearly resized with preserved aspect ratio and centered on a 224$\times$224 gray-127 canvas. All encoders receive ImageNet mean/std normalization. This common event transform differs from the standard supervised ImageNet center-crop evaluation recipe. It is image-like inference on real events, not native asynchronous inference.\par
Every clip supplies a centered 1.6 s interval; none is shorter. The original grid uses \(W\in\{20,30,50\}\) ms and \(N\in\{1,2,4,8,16,32\}\) views. For two or more views, right boundaries span 50--1600 ms uniformly, rounded to microseconds; a single view is the final snapshot. Windows do not overlap and can leave unsampled gaps. More views change retained evidence as well as compute. Centering requires the full clip length and does not implement online onset detection.\par
All new baseline, readout-stability, and calibration comparisons fix 50 ms/eight views. Exact boundaries are 50, 271.429, 492.857, 714.286, 935.714, 1157.143, 1378.571, and 1600 ms. Two original controls additionally vary history (400/800/1600 ms at eight views) or retain 25/50/100\% of real events with three fixed thinning seeds. They modify encoder inputs, not measured sensor energy.\par
\subsection{Encoders and readout}
LeJEPA-S/B contain 21,665,664/85,798,656 encoder parameters, with 12-layer patch-16 ViT-v2 backbones and widths 384/768. DINO-S/B use corresponding publisher configurations. We verify checkpoint hashes, identify excluded training projections, and load encoders strictly. We do not independently establish the DINO export's teacher/student role from a model-card evaluation statement.\par
Patch mean and CLS are alternative frozen readouts. Each view's readout is L2 normalized; the clip feature averages eight view vectors. This aggregation discards order. The V-JEPA 2.1 distilled-B EMA reference (86,833,152 parameters) encodes independent 224-pixel views through its image path with the existing rotary interpolation setting, rather than using temporal attention. It has different pretraining and nominal resolution and is a cross-family reference.\par
MobileNetV3-Small and ResNet-18 use Torchvision 0.24.1 IMAGENET1K\_V1 weights from official hash-checked downloads. We remove their classification layers and spatially average, L2 normalize each view, and average views. Encoder parameter counts are 927,008 and 11,176,512, respectively; full published classification models contain 2,542,856 and 11,689,512. These are frozen supervised-pretrained features, not event-trained encoders.\par
TinyCNN has 61,586 parameters: four stride-2 convolutions (5$\times$5, 3$\rightarrow$16; 3$\times$3, 16$\rightarrow$32$\rightarrow$64$\rightarrow$64), each with four-group GroupNorm and ReLU, spatial averaging, view averaging, and a 64$\rightarrow$2 linear head. Each policy/fold is trained independently from scratch with all three seeds 20261022/23/24. It is a simple trained image-like event baseline, not a state-of-the-art event-native model or an optimized architecture search.\par
\subsection{Fitting and validation}
Each person belongs to one of five folds. Test fold \(f\), validation fold \((f+1)\bmod5\), and the remaining training folds are disjoint in people. Original test clip counts are 111/111/95/97/106; people counts are 10/10/9/9/9. There is one out-of-fold test prediction per clip per setting.\par
For frozen features, standardization is fitted on training clips only, followed by two-output ridge regression with alpha in \{0.01, 0.1, 1, 10, 100\}. The score is positive minus negative output. For each alpha, validation thresholds maximize capture subject to FPR budgets of 5/10/20\%, with ties favoring fewer false triggers and then a higher threshold; always-off is allowed. Mean validation capture across budgets selects alpha, ties favoring stronger regularization. Selected thresholds are then applied unchanged to test clips. The 10\% validation budget does not enforce 10\% test FPR.\par
TinyCNN uses 30 epochs, AdamW (learning rate 0.001, weight decay 0.0001), clip batch eight, training-only class weights, and no augmentation. Each epoch is evaluated only on validation clips. The same three-budget mean capture selects the checkpoint, ties favoring the earliest epoch. A checkpoint and its thresholds are saved before test scoring. All three training seeds are reported without selecting or ensembling them. This training regime is not identical to frozen-feature fitting.\par
\section{Evaluation and Analysis Scope}
\subsection{Development exposure and repeated grouping}
These clips and all 47 participants had prior experimental use. The original 148 input conditions were fixed before their scoring, but were never an untouched confirmation set. The new baselines, repeated grouping, and calibration analyses are explicitly post-review extensions; they do not retroactively become original predeclared hypotheses.\par
Ten new fixed grouping seeds (20261101--20261110) repartition the same participants using target-availability strata and deterministic SHA sorting. Each repeat uses five grouped train/validation/test rotations, the same fitting procedure, and all ten frozen representations (four ViTs $\times$ two readouts plus two CNNs): 1,000 new fitted heads. All repeats are retained. Their spread measures partition sensitivity on known development participants, not independent population replications. TinyCNN is evaluated on the original grouping with its three seeds, rather than on this full repeated grid.\par
\subsection{Metrics and uncertainty}
Capture is TP/(TP+FN); false-trigger rate is FP/(FP+TN), with counts reported. One clip gives one decision. The selected class mixture cannot establish everyday prevalence, precision, recorded duration, or false triggers per hour.\par
ROC curves are computed within each test fold because different fold heads need not share a score scale. We report the unweighted fold mean of AUROC, normalized partial area from FPR 0 to 0.2 (pAUC20 = area/0.2), and linearly interpolated TPR at FPR 0.1 (TPR10). Ties share a ROC step. pAUC20 ranges from zero to one; a random diagonal has 0.1, not 0.5. This is not McClish-standardized partial AUC. Test ROC interpolation is a retrospective matched-FPR diagnostic, not a deployable validation-calibrated threshold or an estimate of calibration reliability. Interpolated points need not be attainable by a deterministic threshold when scores tie.\par
To examine calibration separately, we hold each original head and alpha fixed, bootstrap validation people 1,000 times within folds, and recalibrate the 10\% threshold. Unchanged test scores yield a distribution of aggregate capture/FPR. Its 5th/50th/95th percentiles describe conditional sensitivity; they exclude refitting and pretraining uncertainty and are not confidence limits for generalization.\par
The three original contrasts (scale, duration, view count) under both policies retain their original definitions. A post-review reanalysis applies 50,000 paired participant resamples and Bonferroni percentile intervals across six contrasts $\times$ two metrics (family 12). These conditional, approximate bootstrap intervals do not establish fresh confirmatory significance. Original unadjusted 5,000-resample intervals remain archived, rather than being silently replaced.\par
\subsection{Implementation and cost}
Raw CRC/content hashes, exact views, pinned weights, fold heads, and selections are retained. Cached legacy features are reused only at matching content, times, preprocessing, and weights. Encoder timing uses M1 Pro/16 GB, MPS FP32, four CPU threads, batch one view, ten warm-ups and 50 measured runs on a fixed training input; CPU return synchronizes inference. CNN references were measured in a separate session before CNN training. These are host-implementation costs, not eyewear energy. Timing excludes offline decoding and input preparation unless a stage is named explicitly.\par
\section{Results}
\subsection{Readout, model scale, and lightweight baselines}
\begin{table*}[t]
\caption{P1 at 50 ms/eight views. Capture and false triggers use validation-selected thresholds; pAUC and TPR10 are fold-wise test-ROC diagnostics, not calibrated deployment thresholds.}
\centering\footnotesize
\fitTable{\begin{tabular}{lrrrr}
\toprule
Representation & Targets captured & Non-target triggers & pAUC20 & TPR10 \\
\midrule
LeJEPA-S patch & 78/132 (59.1\%) & 47/388 (12.1\%) & 0.603 & 0.639 \\
LeJEPA-B patch & 89/132 (67.4\%) & 35/388 (9.0\%) & 0.670 & 0.709 \\
DINO-S patch & 78/132 (59.1\%) & 35/388 (9.0\%) & 0.635 & 0.681 \\
DINO-B patch & 88/132 (66.7\%) & 33/388 (8.5\%) & 0.682 & 0.703 \\
LeJEPA-S CLS & 78/132 (59.1\%) & 47/388 (12.1\%) & 0.533 & 0.566 \\
LeJEPA-B CLS & 87/132 (65.9\%) & 44/388 (11.3\%) & 0.657 & 0.682 \\
DINO-S CLS & 101/132 (76.5\%) & 40/388 (10.3\%) & 0.732 & 0.797 \\
DINO-B CLS & 98/132 (74.2\%) & 37/388 (9.5\%) & 0.746 & 0.809 \\
MobileNetV3-Small & 97/132 (73.5\%) & 53/388 (13.7\%) & 0.613 & 0.648 \\
ResNet-18 & 85/132 (64.4\%) & 42/388 (10.8\%) & 0.569 & 0.630 \\
TinyCNN, seed 20261022 & 31/132 (23.5\%) & 54/388 (13.9\%) & 0.142 & 0.143 \\
TinyCNN, seed 20261023 & 34/132 (25.8\%) & 72/388 (18.6\%) & 0.132 & 0.121 \\
TinyCNN, seed 20261024 & 29/132 (22.0\%) & 59/388 (15.2\%) & 0.145 & 0.140 \\
\bottomrule\end{tabular}}
\end{table*}
DINO-S changes from 59.1\% capture/9.0\% false triggers with patch mean to 76.5\%/10.3\% with CLS. Both low-FPR diagnostics also improve (pAUC20 0.635$\rightarrow$0.732; TPR10 0.681$\rightarrow$0.797). LeJEPA-S changes in the opposite direction for pAUC20 (0.603$\rightarrow$0.533), despite identical aggregate selected-threshold TP/FP counts. All 520 clip features differ between its readouts, 58 P1 clip decisions differ, and independent forwards reproduce each cache; the identical counts are not a fallback to patch mean. CLS is therefore not universally preferable, and readout effects cannot be generalized across families from the DINO-S gain.\par
DINO-S CLS has higher selected-threshold capture than DINO-B CLS but also more false triggers. DINO-B has higher pAUC20 and TPR10. This reversal demonstrates why a single calibrated operating point does not rank representations by discrimination. LeJEPA and DINO comparisons establish no JEPA-specific advantage; they also do not prove equivalence.\par
The 0.927M-parameter MobileNet encoder captures 97/132 with 53/388 false triggers, compared with LeJEPA-B's 89/132 and 35/388. It is competitive in capture but is operating at a different FPR; its original-split pAUC20 is lower than both DINO CLS readouts. ResNet-18 is not uniformly stronger despite its larger parameter count. The trained TinyCNN's P1 capture spans 22.0\%--25.8\%, with 13.9\%--18.6\% false triggers across its three seeds. Table 1 reports every seed separately; no best-seed result replaces the others. The limited training recipe and small sample do not establish the best attainable performance of compact CNNs.\par
\subsection{Partition and threshold sensitivity}
\begin{table*}[t]
\caption{Repeated participant grouping: mean [minimum, maximum] normalized pAUC20 over all ten fixed repartitions. These dependent repeats use the same 47 development participants.}
\centering\footnotesize
\fitTable{\begin{tabular}{lrr}
\toprule
Representation & P1 pAUC20 & P2 pAUC20 \\
\midrule
LeJEPA-S patch & 0.580 [0.554, 0.635] & 0.532 [0.483, 0.585] \\
LeJEPA-B patch & 0.681 [0.626, 0.727] & 0.654 [0.591, 0.718] \\
DINO-S patch & 0.621 [0.587, 0.668] & 0.574 [0.528, 0.636] \\
DINO-B patch & 0.658 [0.625, 0.690] & 0.641 [0.591, 0.676] \\
LeJEPA-S CLS & 0.542 [0.491, 0.602] & 0.531 [0.477, 0.582] \\
LeJEPA-B CLS & 0.649 [0.603, 0.696] & 0.632 [0.587, 0.680] \\
DINO-S CLS & 0.696 [0.604, 0.751] & 0.619 [0.573, 0.666] \\
DINO-B CLS & 0.738 [0.686, 0.779] & 0.706 [0.651, 0.747] \\
MobileNetV3-Small & 0.674 [0.631, 0.715] & 0.617 [0.540, 0.655] \\
ResNet-18 & 0.592 [0.515, 0.633] & 0.556 [0.506, 0.614] \\
\bottomrule\end{tabular}}
\end{table*}
Across ten repartitions, DINO-B CLS has mean P1 pAUC20 0.738, DINO-S CLS 0.696, MobileNet 0.674, and LeJEPA-B patch mean 0.681. MobileNet's original pAUC20 of 0.613 is below its repeated mean, illustrating that conclusions from one grouping can move. The repeated means summarize the fixed set of partitions; overlapping data prevent treating ten means as ten independent experiments. P2 is retained rather than selecting the more favorable policy. Within DINO, changing patch mean to CLS increases mean P1 pAUC20 by 0.075 for S and 0.080 for B across these repartitions, whereas S$\rightarrow$B at patch mean increases it by 0.037. Both readout changes are positive in all ten partitions. This is a within-family descriptive comparison, not a universal dominance of readout over size: LeJEPA patch-mean S$\rightarrow$B increases mean pAUC20 by 0.101, while its S readout change averages -0.038. The same participants are reused throughout.\par
\begin{figure*}[t]
\centering\includegraphics[width=\textwidth,height=.34\textheight,keepaspectratio]{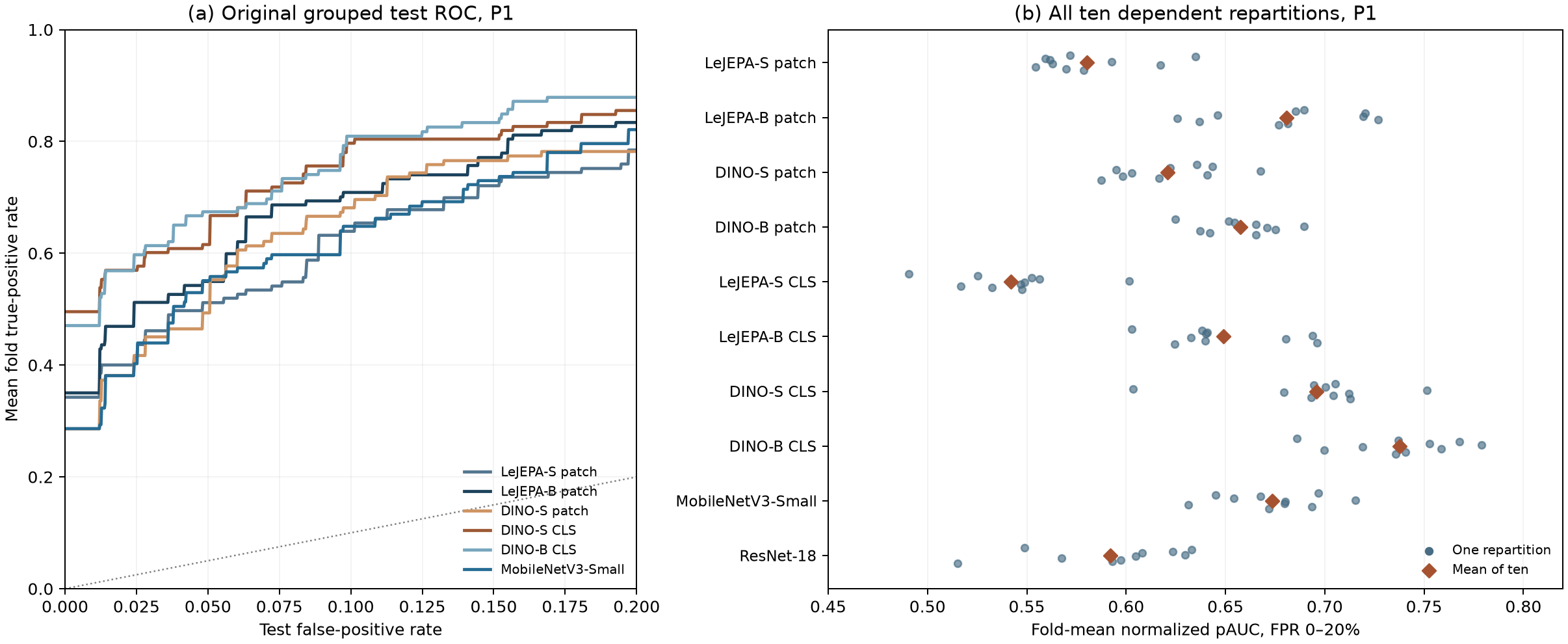}
\caption{Readout and representation sensitivity: original fold-wise P1 ROC and all ten repeated-group pAUC20 values. Points are dependent repartitions, not independent confidence samples.}
\end{figure*}
\begin{table*}[t]
\caption{Validation-threshold sensitivity, P1: 5th / 50th / 95th percentiles (\%) from 1,000 validation-participant bootstrap recalibrations, holding original heads and alpha fixed. These are conditional sensitivity summaries.}
\centering\footnotesize
\fitTable{\begin{tabular}{lrr}
\toprule
Representation & Capture (\%) & False triggers (\%) \\
\midrule
LeJEPA-S patch & 53.8 / 59.1 / 63.6 & 10.1 / 12.1 / 14.7 \\
LeJEPA-B patch & 51.5 / 64.4 / 73.5 & 6.2 / 8.8 / 12.4 \\
DINO-S CLS & 63.6 / 72.0 / 78.0 & 6.2 / 9.8 / 14.2 \\
DINO-B CLS & 68.2 / 72.7 / 77.3 & 6.4 / 9.0 / 12.4 \\
MobileNetV3-Small & 60.6 / 69.7 / 75.8 & 9.3 / 12.4 / 15.7 \\
\bottomrule\end{tabular}}
\end{table*}
LeJEPA-B patch-mean test FPR ranges from 6.2\% to 12.4\% between the 5th and 95th calibration percentiles; MobileNet spans 9.3\%--15.7\%. Only validation participants are resampled. Head scores and alpha stay fixed, so this captures one source of threshold instability rather than total pipeline uncertainty. More calibration data, cross-source backgrounds, and temporal controls are needed before a strict deployment false-alarm budget can be claimed.\par
\subsection{Accumulation budgets and multiplicity}
\begin{table*}[t]
\caption{Primary-policy results at a validation-selected 10\% non-target trigger budget. Test rates are realized rates, not matched-FPR comparisons.}
\centering\footnotesize
\fitTable{\begin{tabular}{lrrrr}
\toprule
Encoder & Window (ms) & Views & Targets captured & Non-targets triggered \\
\midrule
LeJEPA-S & 20 & 8 & 76/132 (57.6\%) & 49/388 (12.6\%) \\
LeJEPA-S & 50 & 8 & 78/132 (59.1\%) & 47/388 (12.1\%) \\
LeJEPA-B & 50 & 8 & 89/132 (67.4\%) & 35/388 (9.0\%) \\
LeJEPA-S & 50 & 32 & 95/132 (72.0\%) & 56/388 (14.4\%) \\
\bottomrule\end{tabular}}
\end{table*}
\begin{table*}[t]
\caption{Original paired contrasts, percentage-point differences with post-review Bonferroni family-12 bootstrap intervals (50,000 participant resamples). All intervals condition on fixed original folds and fitted heads.}
\centering\footnotesize
\fitTable{\begin{tabular}{lrrr}
\toprule
Policy & Change & Capture difference & False-trigger difference \\
\midrule
P1 & S$\rightarrow$B, 50 ms/8 views & +8.3 [-0.8, +19.0] & -3.1 [-8.7, +2.5] \\
P1 & 20$\rightarrow$50 ms, S/8 views & +1.5 [-7.3, +9.8] & -0.5 [-6.9, +5.2] \\
P1 & 8$\rightarrow$32 views, S/50 ms & +12.9 [+2.1, +26.6] & +2.3 [-1.3, +6.8] \\
P2 & S$\rightarrow$B, 50 ms/8 views & +11.2 [+1.3, +21.0] & -1.8 [-6.6, +2.8] \\
P2 & 20$\rightarrow$50 ms, S/8 views & +8.4 [+0.5, +17.3] & +0.9 [-4.7, +6.7] \\
P2 & 8$\rightarrow$32 views, S/50 ms & +9.0 [-0.6, +19.8] & +1.2 [-3.6, +6.1] \\
\bottomrule\end{tabular}}
\end{table*}
LeJEPA's original P1 S$\rightarrow$B capture difference is +8.3 points. Its family-adjusted interval is [-0.8, +19.0], spanning zero; the original marginal interval should not be read as robust multiple-comparison evidence. The 8$\rightarrow$32-view P1 capture difference remains positive in its adjusted interval, but its actual false-trigger rate also increases and inputs are not held constant. P2 differs in its duration response. These are conditional development-data observations, not a general scaling law or proof of an optimal 20--50 ms range.\par
The full original grid remains in the supplement and machine-readable export. Gains are not universally monotone: LeJEPA-B at 50 ms changes from 71.2\% capture/9.8\% false triggers at 16 views to 70.5\%/10.8\% at 32. This observation does not identify blur or feature dilution as the cause. With eight 50 ms views, LeJEPA-B capture rises from 54.5\% at 400 ms available history to 67.4\% at 1600 ms, while DINO-B is higher at 800 ms (68.2\%) than 1600 ms (66.7\%). History also changes sample locations. Half-event retention gives LeJEPA-S mean capture 58.1\%/false triggers 11.3\% across three seeds, versus 59.1\%/12.1\% at full retention. Independently refitted retention conditions measure input-conditioned sensitivity, not robustness of one fixed deployed head or reduced sensor power.\par
\subsection{Cost and evidence limits}
\begin{table*}[t]
\caption{Local warm M1 Pro/MPS FP32 median timings. One-view timing includes encoding, normalization of the embedding and CPU return. Recompute includes the classifier, but excludes event rendering and image normalization. CNN timings are a separate-session reference.}
\centering\footnotesize
\fitTable{\begin{tabular}{lrrrr}
\toprule
Encoder & Encoder parameters & One view (ms) & 8-view recompute (ms) & 32-view recompute (ms) \\
\midrule
LeJEPA-S & 21,665,664 & 18.54 & 146.47 & 413.98 \\
LeJEPA-B & 85,798,656 & 22.96 & 196.09 & 768.16 \\
MobileNetV3-Small & 927,008 & 10.65 & not measured & not measured \\
ResNet-18 & 11,176,512 & 8.23 & not measured & not measured \\
\bottomrule\end{tabular}}
\end{table*}
Parameter count does not predict this host's runtime ordering: MobileNet is smaller but slower per view than ResNet-18 in the measured setup. Cached aggregation/classification at 32 views takes about 0.014/0.024 ms for LeJEPA S/B, excluding new-view encoding. A streaming implementation need not recompute every historical embedding. Separate CPU medians for event accumulation (about 0.83--0.93 ms) and resize/pad (about 0.44 ms) must not be summed and called measured end-to-end latency. No timing establishes eye-worn power or thermal feasibility.\par
\section{Discussion and Limitations}
The practical finding is an evaluation dependency: readout and calibrated working points can change the apparent scale comparison. In these releases, DINO CLS is a strong readout, while LeJEPA-S CLS weakens low-FPR discrimination. A compact supervised encoder remains competitive. Any recommendation to compress a particular large JEPA model should first establish its advantage against such controls at an appropriate operating point.\par
This benchmark addresses an action-defined offline candidate, not subjective recording value. It uses a selected eight-class, fixed-camera third-person subset with centered, action-containing clips and prior development exposure. Participant isolation addresses identity overlap inside a rotation but does not undo dataset adaptation. Repartitioning the same people cannot establish an independent-source result. Mean pooling also omits temporal order, and we have not compared full video inference or a strong event-native model.\par
The motivating wearable system remains unvalidated. There are no continuous idle backgrounds, annotated online trigger deadlines, synchronized IMU/audio, RGB wake-up measurements, or device power measurements. Concatenating action clips would impose artificial event prevalence and boundaries; we do not report it as real continuous false alarms per hour. Personal-interest labels, wearer motion, cross-source backgrounds, and hardware measurements are concrete next stages, rather than outcomes of this study.\par
\section{Conclusion}
Real event accumulations support useful offline candidate discrimination, but the apparent benefit of model size depends on readout and operating-point reporting. Lightweight baselines, low-FPR curves, repeated participant grouping, and calibration sensitivity materially change the interpretation of one capture score. Our original accumulation-budget results remain policy dependent, with input evidence and compute coupled. The release preserves both original and post-review analyses and their uncertainty boundaries. These findings motivate better-controlled evaluation before extrapolating to personal memory triggers or all-day eyewear.\par
\section*{References}
\noindent [1] G. Gallego et al. ``Event-based Vision: A Survey.'' IEEE TPAMI, arXiv:1904.08405v3, 2020. \href{https://arxiv.org/abs/1904.08405v3}{Source}.\par
\noindent [2] P. Bonazzi, J. Moosmann, A. Celik, P. Mayer, and M. Magno. ``OpenGlass: Ultra-Low-Power On-Device AI Eyewear with Event-based Vision.'' arXiv:2606.07431v2, 2026. \href{https://arxiv.org/abs/2606.07431v2}{Source}.\par
\noindent [3] P. Bhattacharyya et al. ``Helios: An extremely low power event-based gesture recognition for always-on smart eyewear.'' arXiv:2407.05206v4, 2024. \href{https://arxiv.org/abs/2407.05206v4}{Source}.\par
\noindent [4] Q. Wang et al. ``DailyDVS-200: A Comprehensive Benchmark Dataset for Event-Based Action Recognition.'' ECCV, 2024. \href{https://arxiv.org/abs/2407.05106v2}{Source}.\par
\noindent [5] R. Fan and W. Hao. ``SMV-EAR: Bring Spatiotemporal Multi-View Representation Learning into Efficient Event-Based Action Recognition.'' arXiv:2601.17391v1, 2026. \href{https://arxiv.org/abs/2601.17391v1}{Source}.\par
\noindent [6] M. Assran et al. ``Self-Supervised Learning From Images With a Joint-Embedding Predictive Architecture.'' CVPR, pp. 15619--15629, 2023. \href{https://arxiv.org/abs/2301.08243}{Source}.\par
\noindent [7] A. Bardes et al. ``Revisiting Feature Prediction for Learning Visual Representations from Video.'' arXiv:2404.08471, 2024. \href{https://arxiv.org/abs/2404.08471}{Source}.\par
\noindent [8] R. Balestriero and Y. LeCun. ``LeJEPA: Provable and Scalable Self-Supervised Learning Without the Heuristics.'' arXiv:2511.08544v3, 2025. \href{https://arxiv.org/abs/2511.08544v3}{Source}.\par
\noindent [9] M. Caron et al. ``Emerging Properties in Self-Supervised Vision Transformers.'' arXiv:2104.14294v2, 2021. \href{https://arxiv.org/abs/2104.14294v2}{Source}.\par
\noindent [10] Open-Knowledge-AI. LeJEPA ViT-S/16 ImageNet-1K release, revision cc7022877d51494709ef398d437fb8619349e0f9. \href{https://huggingface.co/OK-AI/lejepa-vits16-pretrain-in1k/tree/cc7022877d51494709ef398d437fb8619349e0f9}{Pinned release}.\par
\noindent [11] Open-Knowledge-AI. LeJEPA ViT-B/16 ImageNet-1K release, revision 5dd71e2fc9a8e97d84c047f3a8cd5bf19ff199fa. \href{https://huggingface.co/OK-AI/lejepa-vitb16-pretrain-in1k/tree/5dd71e2fc9a8e97d84c047f3a8cd5bf19ff199fa}{Pinned release}.\par
\noindent [12] Open-Knowledge-AI. DINO ViT-S/16 ImageNet-1K release, revision fa32cfe8bd41d44462e99336601d9f1ec5bf6ccc. \href{https://huggingface.co/OK-AI/dino-vits16-pretrain-in1k/tree/fa32cfe8bd41d44462e99336601d9f1ec5bf6ccc}{Pinned release}.\par
\noindent [13] Open-Knowledge-AI. DINO ViT-B/16 ImageNet-1K release, revision d005b9ab18098bc8af7688f9b66affbe73019bc9. \href{https://huggingface.co/OK-AI/dino-vitb16-pretrain-in1k/tree/d005b9ab18098bc8af7688f9b66affbe73019bc9}{Pinned release}.\par
\noindent [14] L. Mur-Labadia et al. ``V-JEPA 2.1: Unlocking Dense Features in Video Self-Supervised Learning.'' arXiv:2603.14482v3, 2026. \href{https://arxiv.org/abs/2603.14482v3}{Source}.\par
\noindent [15] A. Howard et al. ``Searching for MobileNetV3.'' ICCV, 2019. \href{https://arxiv.org/abs/1905.02244v5}{Source}.\par
\noindent [16] K. He, X. Zhang, S. Ren, and J. Sun. ``Deep Residual Learning for Image Recognition.'' CVPR, 2016. \href{https://arxiv.org/abs/1512.03385}{Source}.\par
\section*{Reproducibility}
The accompanying local release candidate includes the original 148-condition/888-operating-point export, per-clip outputs, the primary frozen features, standalone downstream refitting and verification code, grouped splits, model provenance, training recipes, and post-review analyses. The original matrix comprises 296 policy-condition pairs and 1,480 fold-specific heads; new CNN baselines and repeated grouping are separate additions. Raw DailyDVS recordings and third-party weights are obtained under their source terms and are not redistributed. The release is prepared locally; no public repository or preprint identifier is claimed.\par
\section*{AI assistance}
AI tools assisted implementation, experiment orchestration, checking, and drafting. Binary targets are deterministic mappings of dataset labels. Reported scores are produced by the documented encoders and supervised classifiers, not language-model semantic judgments. The author is responsible for the content.\par

\clearpage\onecolumn\raggedbottom
\setcounter{table}{0}\renewcommand{\thetable}{S\arabic{table}}
\renewcommand{\theHtable}{supp.\arabic{table}}
\setcounter{figure}{0}\renewcommand{\thefigure}{S\arabic{figure}}
\renewcommand{\theHfigure}{supp.\arabic{figure}}
\section*{Supplementary Material}

This supplement separates the original fixed evaluation from explicitly post-review extensions. Rates describe offline clip decisions, not recording value, online trigger frequency, or device energy.\par
\section*{S1. Data and grouped splits}
\begin{table}[H]
\caption{Participant-disjoint fold composition.}
\centering\footnotesize
\fitTable{\begin{tabular}{lrrrrr}
\toprule
Test fold & Test people & Test clips & P1 targets & P2 targets & Validation fold \\
\midrule
0 & 10 & 111 & 28 & 38 & 1 \\
1 & 10 & 111 & 28 & 38 & 2 \\
2 & 9 & 95 & 24 & 32 & 3 \\
3 & 9 & 97 & 25 & 33 & 4 \\
4 & 9 & 106 & 27 & 37 & 0 \\
\bottomrule\end{tabular}}
\end{table}
Remaining three folds form the training group. No participant crosses the training, validation and test groups within a fold. All people have prior research exposure; these folds are not a fresh confirmation set.\par
\section*{S2. Complete primary-setting budget points}
Each budget is a validation constraint. The realized test non-target rate need not satisfy it. Each entry is capture / non-target trigger rate (\%).\par
\begin{table}[H]
\caption{Primary-setting validation budgets: capture / realized false triggers.}
\centering\footnotesize
\fitTable{\begin{tabular}{lrrrr}
\toprule
Encoder (50 ms, 8 views) & Policy & 5\% validation budget & 10\% validation budget & 20\% validation budget \\
\midrule
lejepa\_s & P1 & 52.3 / 9.8 & 59.1 / 12.1 & 68.9 / 19.8 \\
lejepa\_s & P2 & 51.7 / 8.8 & 57.9 / 11.4 & 67.4 / 17.8 \\
lejepa\_b & P1 & 50.8 / 4.9 & 67.4 / 9.0 & 78.8 / 16.8 \\
lejepa\_b & P2 & 59.6 / 5.6 & 69.1 / 9.6 & 80.3 / 17.5 \\
dino\_s & P1 & 41.7 / 3.6 & 59.1 / 9.0 & 71.2 / 18.3 \\
dino\_s & P2 & 42.1 / 5.3 & 56.7 / 8.8 & 71.3 / 20.2 \\
dino\_b & P1 & 54.5 / 2.8 & 66.7 / 8.5 & 84.1 / 20.9 \\
dino\_b & P2 & 58.4 / 4.1 & 66.3 / 8.8 & 79.8 / 21.3 \\
vjepa\_b & P1 & 55.3 / 4.6 & 63.6 / 7.0 & 72.7 / 14.7 \\
vjepa\_b & P2 & 62.4 / 11.4 & 67.4 / 14.3 & 72.5 / 17.3 \\
\bottomrule\end{tabular}}
\end{table}
\section*{S3. Class-level errors at the predeclared primary setting}
P1 target classes are handshake (71) and fall (109); every other class is a non-target by policy definition. Reported counts are selected clips divided by clips available in that class. This is recall for target classes and false triggering for non-target classes.\par
\begin{table}[H]
\caption{Class-level selections under P1.}
\centering\footnotesize
\fitTable{\begin{tabular}{lrrrrr}
\toprule
Action (class ID) & Clips & LeJEPA-S selected & LeJEPA-B selected & DINO-S selected & DINO-B selected \\
\midrule
Drink (1) & 70 & 8 & 7 & 4 & 5 \\
Take from bag (8) & 70 & 5 & 4 & 4 & 5 \\
Phone use (31) & 70 & 4 & 5 & 9 & 5 \\
Put into bag (62) & 70 & 9 & 6 & 6 & 7 \\
Handshake (target) (71) & 70 & 40 & 50 & 46 & 50 \\
Pick up object (100) & 62 & 18 & 12 & 9 & 11 \\
Fall (target) (109) & 62 & 38 & 39 & 32 & 38 \\
Open door (158) & 46 & 3 & 1 & 3 & 0 \\
\bottomrule\end{tabular}}
\end{table}
The picking-up action is retained as a non-target for these fixed policies, rather than excluded after errors are observed. Class counts are descriptive and were not used to change the target policy or select hyperparameters.\par
\section*{S4. V-JEPA cross-family reference and CLS sensitivity}
V-JEPA denotes the official V-JEPA 2.1 distilled ViT-B EMA encoder (86,833,152 parameters), evaluated at 224 pixels using its image path: one temporal image dimension, independent per-view encoding, mean patch tokens, per-view L2 normalization, and mean aggregation. Its checkpoint has nominal 384-pixel configuration; this reference uses the same 224-pixel input size as the main study. It is neither a full temporal-video evaluation nor a controlled objective-only comparison. CLS readout is a predeclared sensitivity check, not a replacement selected from test outcomes.\par
\begin{table}[H]
\caption{V-JEPA image-path reference and CLS sensitivity.}
\centering\footnotesize
\fitTable{\begin{tabular}{lrrrrr}
\toprule
Encoder & Readout & W (ms) & Views & P1 capture / false triggers (\%) & P2 capture / false triggers (\%) \\
\midrule
vjepa\_b & patch\_mean & 20 & 1 & 53.8 / 8.8 & 49.4 / 9.9 \\
vjepa\_b & patch\_mean & 20 & 8 & 60.6 / 8.2 & 68.0 / 11.7 \\
vjepa\_b & patch\_mean & 20 & 32 & 66.7 / 9.3 & 68.5 / 15.2 \\
vjepa\_b & patch\_mean & 50 & 1 & 49.2 / 9.8 & 38.2 / 9.6 \\
vjepa\_b & patch\_mean & 50 & 8 & 63.6 / 7.0 & 67.4 / 14.3 \\
vjepa\_b & patch\_mean & 50 & 32 & 70.5 / 10.6 & 67.4 / 12.9 \\
lejepa\_s & cls & 50 & 8 & 59.1 / 12.1 & 53.4 / 9.9 \\
lejepa\_b & cls & 50 & 8 & 65.9 / 11.3 & 62.9 / 8.5 \\
dino\_s & cls & 50 & 8 & 76.5 / 10.3 & 66.3 / 10.2 \\
dino\_b & cls & 50 & 8 & 74.2 / 9.5 & 73.6 / 9.1 \\
\bottomrule\end{tabular}}
\end{table}
\section*{S5. Full grid and audit scope}
The accompanying complete\_results.en.csv contains 888 budgeted operating points: 148 distinct input conditions, two target policies, and three validation budgets. These are frozen-encoder evaluations with 1,480 fitted fold-specific classifier heads, not 148 pretrained-model training runs. The original matrix contains 100 main-model inputs and 48 auxiliary inputs; every condition is present.\par
The full grid includes native 20/30/50 ms accumulations, 1/2/4/8/16/32 views, 400/800/1600 ms available histories, fixed-seed 25/50\% real-event retention, simple controls, train-only PCA128, V-JEPA reference, and CLS checks. Retention conditions fit their own classifiers and thresholds; they do not measure corruption of a fixed deployed classifier.\par
\begin{figure}[H]
\centering\includegraphics[width=\textwidth,height=.34\textheight,keepaspectratio]{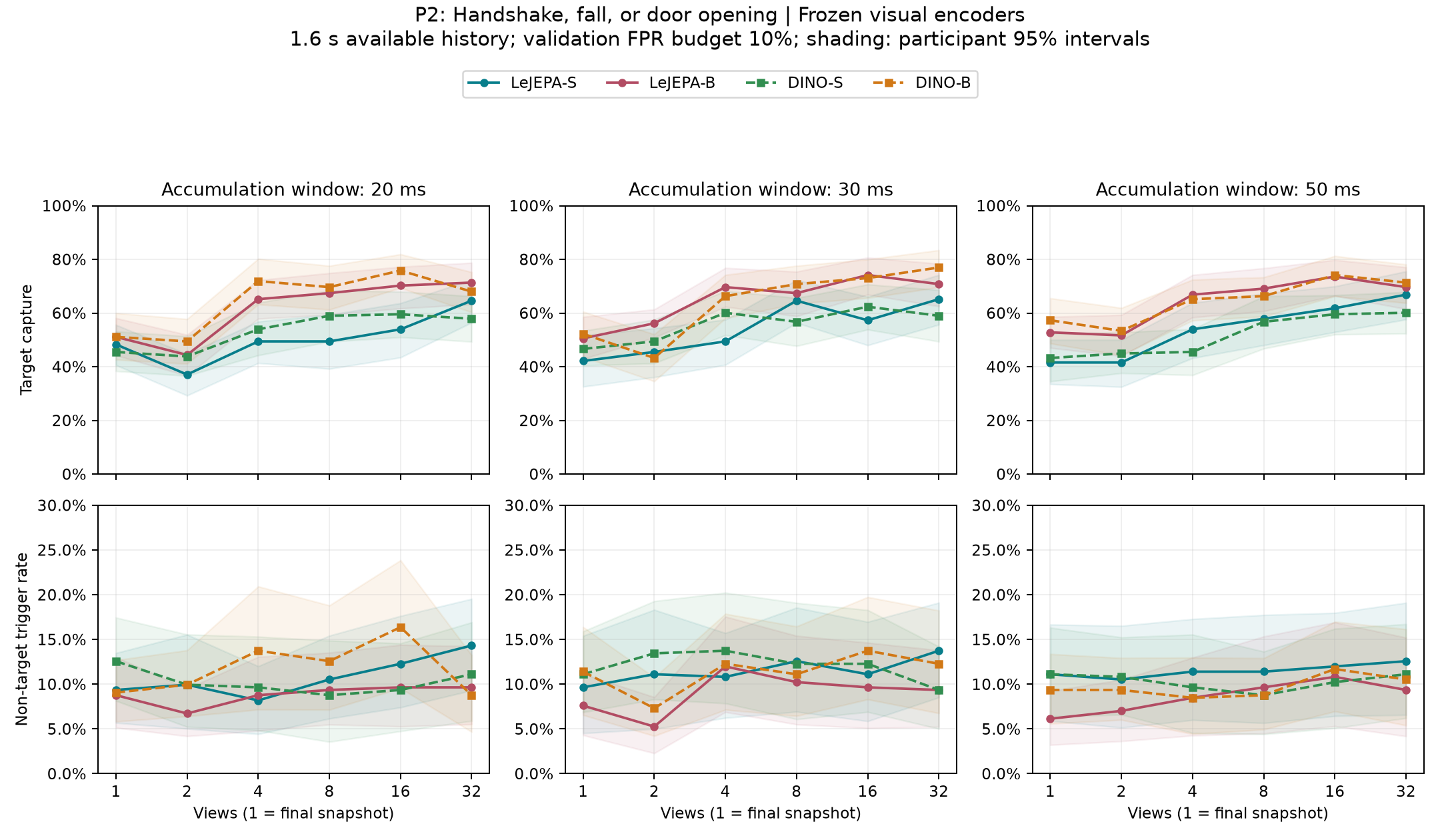}
\caption{Full P2 budget curves.}
\end{figure}
\subsection*{S5.1. Original simple and dimension controls}
The count baseline averages per-window \(\log(1+\text{event count})\), yielding one feature. A 131-dimensional baseline combines that count, positive-polarity proportion, occupied-pixel proportion, and normalized 8$\times$8 spatial counts for each polarity, followed by view averaging. Both baselines use all 18 core input configurations and the same downstream selection procedure.\par
A dimension control fits PCA to the raw aggregated training embeddings only, retains 128 components, and subsequently fits the same standardized ridge classifier. Validation and test participants are not used to fit PCA. This controls downstream dimension, but also changes feature geometry and is not a complete isolation of encoder capacity.\par
\begin{table}[H]
\caption{Original simple-statistic and dimension controls.}
\centering\footnotesize
\fitTable{\begin{tabular}{lrrrr}
\toprule
Representation & P1 capture & P1 false triggers & P2 capture & P2 false triggers \\
\midrule
Event count, 1-D & 12.1\% & 10.3\% & 8.4\% & 10.2\% \\
Event statistics, 131-D & 25.0\% & 14.7\% & 20.2\% & 11.1\% \\
LeJEPA-S & 59.1\% & 12.1\% & 57.9\% & 11.4\% \\
LeJEPA-B & 67.4\% & 9.0\% & 69.1\% & 9.6\% \\
DINO-S & 59.1\% & 9.0\% & 56.7\% & 8.8\% \\
DINO-B & 66.7\% & 8.5\% & 66.3\% & 8.8\% \\
LeJEPA-S, PCA128 & 50.8\% & 8.5\% & 50.6\% & 12.9\% \\
LeJEPA-B, PCA128 & 60.6\% & 14.2\% & 59.6\% & 9.9\% \\
\bottomrule\end{tabular}}
\end{table}
\section*{S6. Original history and event-retention controls}
\begin{table}[H]
\caption{Fixed-eight-view available-history control.}
\centering\footnotesize
\fitTable{\begin{tabular}{lrrr}
\toprule
Encoder & 400 ms history & 800 ms history & 1600 ms history \\
\midrule
LeJEPA-S & 60.6\% / 12.9\% & 56.8\% / 10.8\% & 59.1\% / 12.1\% \\
LeJEPA-B & 54.5\% / 10.1\% & 60.6\% / 9.0\% & 67.4\% / 9.0\% \\
DINO-S & 54.5\% / 8.5\% & 62.9\% / 9.5\% & 59.1\% / 9.0\% \\
DINO-B & 54.5\% / 9.3\% & 68.2\% / 9.3\% & 66.7\% / 8.5\% \\
\bottomrule\end{tabular}}
\end{table}
At 50 ms, LeJEPA-B capture increases from 54.5\% at 400 ms history to 67.4\% at 1600 ms, whereas DINO-B reaches 68.2\% at 800 ms and 66.7\% at 1600 ms. These point estimates do not define an optimal horizon or a statistically established decrease. The complete control includes both 20 and 50 ms windows for all four encoders. Changing available history also changes temporal sampling positions, even at fixed view count. These are offline interval-prefix comparisons, not measured delays from action onset.\par
\begin{figure}[H]
\centering\includegraphics[width=\textwidth,height=.34\textheight,keepaspectratio]{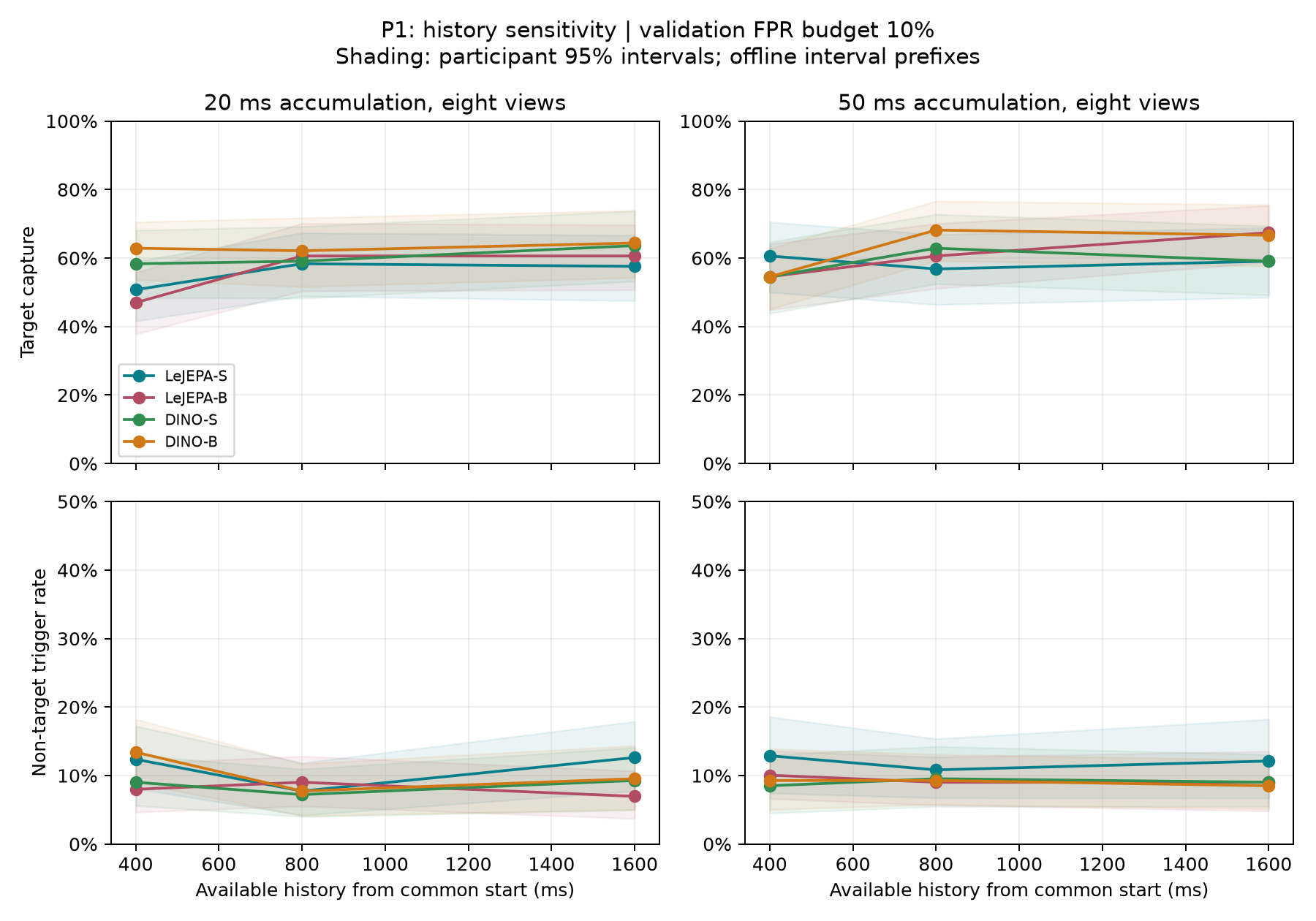}
\caption{Available-history sensitivity at fixed view count.}
\end{figure}
\begin{table}[H]
\caption{Actual-event retention with all thinning seeds.}
\centering\footnotesize
\fitTable{\begin{tabular}{lrrr}
\toprule
Encoder & Retention & Capture: mean [range] & Non-target triggers: mean [range] \\
\midrule
LeJEPA-S & 25\% & 52.5\% [50.8\%, 54.5\%] & 10.7\% [10.3\%, 10.8\%] \\
LeJEPA-S & 50\% & 58.1\% [56.1\%, 60.6\%] & 11.3\% [8.8\%, 13.4\%] \\
LeJEPA-S & 100\% & 59.1\% [59.1\%, 59.1\%] & 12.1\% [12.1\%, 12.1\%] \\
LeJEPA-B & 25\% & 60.1\% [59.8\%, 60.6\%] & 8.9\% [8.8\%, 9.0\%] \\
LeJEPA-B & 50\% & 63.9\% [57.6\%, 68.9\%] & 9.5\% [9.0\%, 10.1\%] \\
LeJEPA-B & 100\% & 67.4\% [67.4\%, 67.4\%] & 9.0\% [9.0\%, 9.0\%] \\
\bottomrule\end{tabular}}
\end{table}
Half retention gives LeJEPA-S a mean capture of 58.1\% and mean non-target trigger rate of 11.3\%, compared with 59.1\% and 12.1\% for all events. LeJEPA-B half-retention capture ranges from 57.6\% to 68.9\% across seeds, cautioning against a single favorable draw. These descriptive ranges do not establish equivalence with full retention. Each retention/seed setting fits its own training-only head and validation threshold. This measures input-conditioned pipeline sensitivity, not test corruption of a fixed classifier. The reduced fractions retain floor(r*n) real events per original window. Three seeds are retained without selecting the best or averaging their predictions into a more expensive ensemble. Results for P2 and all 5/20\% validation budgets are preserved in the full export.\par
\begin{figure}[H]
\centering\includegraphics[width=\textwidth,height=.34\textheight,keepaspectratio]{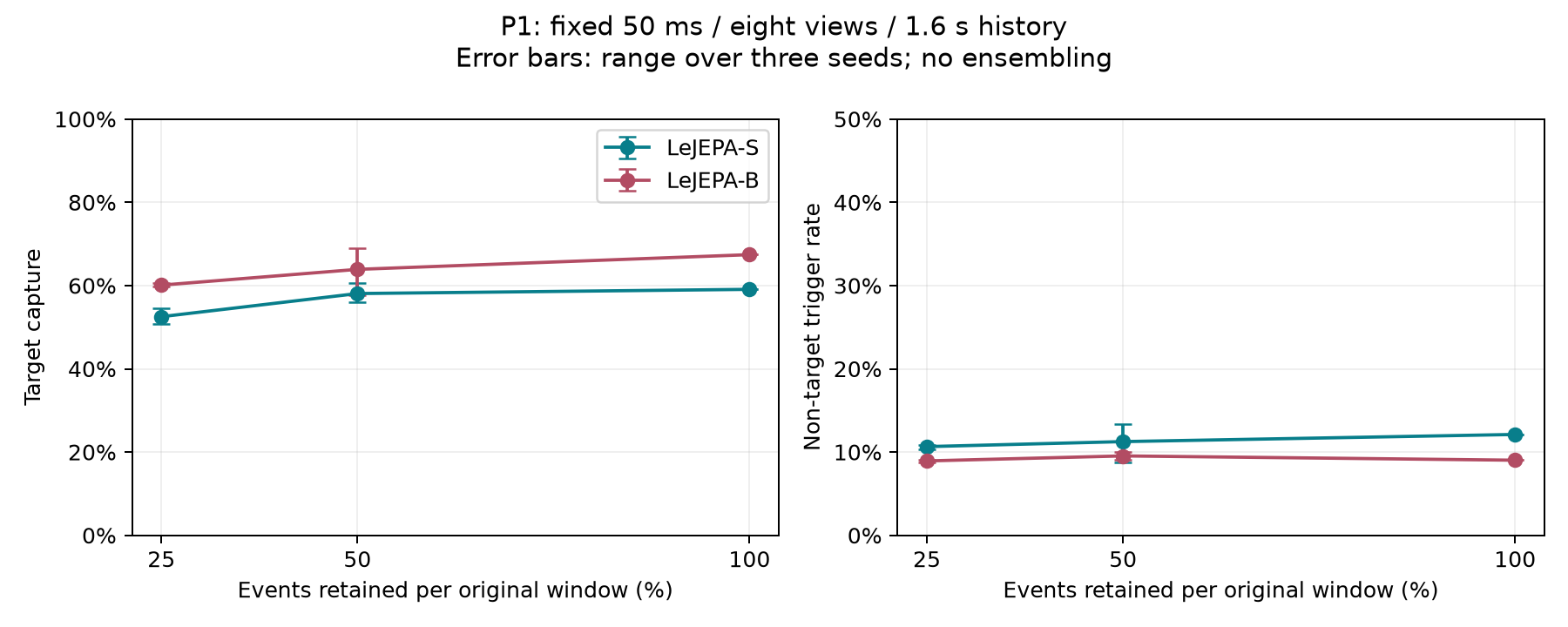}
\caption{Actual-event retention sensitivity.}
\end{figure}
\section*{S7. Complete new baseline and readout operating points}
\begin{table}[H]
\caption{Original grouping: all new baselines and readouts.}
\centering\footnotesize
\fitTable{\begin{tabular}{lrrrrrr}
\toprule
Representation & Policy & 5\% budget & 10\% budget & 20\% budget & pAUC20 & TPR10 \\
\midrule
LeJEPA-S patch & P1 & 52.3\% / 9.8\% & 59.1\% / 12.1\% & 68.9\% / 19.8\% & 0.603 & 0.639 \\
LeJEPA-S patch & P2 & 51.7\% / 8.8\% & 57.9\% / 11.4\% & 67.4\% / 17.8\% & 0.555 & 0.601 \\
LeJEPA-B patch & P1 & 50.8\% / 4.9\% & 67.4\% / 9.0\% & 78.8\% / 16.8\% & 0.670 & 0.709 \\
LeJEPA-B patch & P2 & 59.6\% / 5.6\% & 69.1\% / 9.6\% & 80.3\% / 17.5\% & 0.687 & 0.713 \\
DINO-S patch & P1 & 41.7\% / 3.6\% & 59.1\% / 9.0\% & 71.2\% / 18.3\% & 0.635 & 0.681 \\
DINO-S patch & P2 & 42.1\% / 5.3\% & 56.7\% / 8.8\% & 71.3\% / 20.2\% & 0.582 & 0.615 \\
DINO-B patch & P1 & 54.5\% / 2.8\% & 66.7\% / 8.5\% & 84.1\% / 20.9\% & 0.682 & 0.703 \\
DINO-B patch & P2 & 58.4\% / 4.1\% & 66.3\% / 8.8\% & 79.8\% / 21.3\% & 0.677 & 0.725 \\
LeJEPA-S CLS & P1 & 35.6\% / 4.9\% & 59.1\% / 12.1\% & 72.0\% / 23.5\% & 0.533 & 0.566 \\
LeJEPA-S CLS & P2 & 43.3\% / 4.4\% & 53.4\% / 9.9\% & 65.2\% / 16.4\% & 0.552 & 0.592 \\
LeJEPA-B CLS & P1 & 51.5\% / 4.9\% & 65.9\% / 11.3\% & 78.8\% / 18.8\% & 0.657 & 0.682 \\
LeJEPA-B CLS & P2 & 47.8\% / 6.1\% & 62.9\% / 8.5\% & 72.5\% / 12.9\% & 0.671 & 0.689 \\
DINO-S CLS & P1 & 58.3\% / 1.8\% & 76.5\% / 10.3\% & 81.1\% / 16.0\% & 0.732 & 0.797 \\
DINO-S CLS & P2 & 51.7\% / 4.4\% & 66.3\% / 10.2\% & 77.5\% / 19.3\% & 0.650 & 0.675 \\
DINO-B CLS & P1 & 67.4\% / 4.6\% & 74.2\% / 9.5\% & 85.6\% / 17.3\% & 0.746 & 0.809 \\
DINO-B CLS & P2 & 65.7\% / 7.3\% & 73.6\% / 9.1\% & 83.7\% / 21.9\% & 0.720 & 0.736 \\
MobileNetV3-Small & P1 & 59.1\% / 8.5\% & 73.5\% / 13.7\% & 81.1\% / 19.6\% & 0.613 & 0.648 \\
MobileNetV3-Small & P2 & 55.1\% / 5.0\% & 65.7\% / 11.7\% & 81.5\% / 21.9\% & 0.626 & 0.718 \\
ResNet-18 & P1 & 47.7\% / 5.4\% & 64.4\% / 10.8\% & 75.0\% / 16.8\% & 0.569 & 0.630 \\
ResNet-18 & P2 & 49.4\% / 8.5\% & 59.0\% / 12.0\% & 66.9\% / 17.8\% & 0.528 & 0.563 \\
TinyCNN 20261022 & P1 & 11.4\% / 7.0\% & 23.5\% / 13.9\% & 33.3\% / 24.7\% & 0.142 & 0.143 \\
TinyCNN 20261022 & P2 & 7.9\% / 4.7\% & 21.9\% / 13.2\% & 28.7\% / 17.0\% & 0.164 & 0.191 \\
TinyCNN 20261023 & P1 & 13.6\% / 10.3\% & 25.8\% / 18.6\% & 35.6\% / 25.8\% & 0.132 & 0.121 \\
TinyCNN 20261023 & P2 & 11.2\% / 9.6\% & 26.4\% / 19.0\% & 39.9\% / 29.2\% & 0.129 & 0.123 \\
TinyCNN 20261024 & P1 & 9.1\% / 7.7\% & 22.0\% / 15.2\% & 29.5\% / 21.4\% & 0.145 & 0.140 \\
TinyCNN 20261024 & P2 & 12.9\% / 9.9\% & 23.6\% / 13.2\% & 41.0\% / 24.0\% & 0.163 & 0.147 \\
\bottomrule\end{tabular}}
\end{table}
\section*{S8. Full threshold-recalibration sensitivity}
\begin{table}[H]
\caption{Conditional validation-threshold sensitivity.}
\centering\footnotesize
\fitTable{\begin{tabular}{lrrr}
\toprule
Representation & Policy & Capture & False triggers \\
\midrule
LeJEPA-S patch & P1 & 53.8 / 59.1 / 63.6 & 10.1 / 12.1 / 14.7 \\
LeJEPA-S patch & P2 & 49.4 / 57.9 / 65.2 & 9.1 / 12.3 / 17.0 \\
LeJEPA-B patch & P1 & 51.5 / 64.4 / 73.5 & 6.2 / 8.8 / 12.4 \\
LeJEPA-B patch & P2 & 60.1 / 66.9 / 72.5 & 5.8 / 9.1 / 13.5 \\
DINO-S patch & P1 & 44.7 / 55.7 / 65.2 & 5.2 / 8.5 / 15.2 \\
DINO-S patch & P2 & 45.5 / 54.5 / 63.5 & 6.7 / 10.2 / 16.7 \\
DINO-B patch & P1 & 57.6 / 67.4 / 75.8 & 4.9 / 8.8 / 13.4 \\
DINO-B patch & P2 & 58.4 / 66.3 / 73.6 & 5.6 / 9.1 / 15.5 \\
LeJEPA-S CLS & P1 & 43.1 / 55.3 / 66.7 & 7.0 / 12.4 / 17.8 \\
LeJEPA-S CLS & P2 & 43.2 / 52.8 / 61.8 & 5.0 / 10.5 / 18.7 \\
LeJEPA-B CLS & P1 & 53.0 / 63.6 / 74.2 & 5.9 / 10.3 / 14.7 \\
LeJEPA-B CLS & P2 & 55.1 / 62.9 / 71.3 & 6.1 / 9.1 / 13.2 \\
DINO-S CLS & P1 & 63.6 / 72.0 / 78.0 & 6.2 / 9.8 / 14.2 \\
DINO-S CLS & P2 & 54.5 / 63.5 / 71.9 & 5.3 / 9.9 / 15.2 \\
DINO-B CLS & P1 & 68.2 / 72.7 / 77.3 & 6.4 / 9.0 / 12.4 \\
DINO-B CLS & P2 & 67.4 / 74.2 / 79.8 & 7.6 / 10.8 / 17.3 \\
MobileNetV3-Small & P1 & 60.6 / 69.7 / 75.8 & 9.3 / 12.4 / 15.7 \\
MobileNetV3-Small & P2 & 55.6 / 65.7 / 73.6 & 7.0 / 11.7 / 17.0 \\
ResNet-18 & P1 & 51.5 / 61.4 / 68.2 & 7.7 / 10.3 / 12.9 \\
ResNet-18 & P2 & 47.8 / 56.7 / 62.9 & 8.5 / 11.7 / 14.9 \\
\bottomrule\end{tabular}}
\end{table}
\section*{S9. Training, reproduction, and scope}
All TinyCNN seeds, fold-specific validation-selected epochs, class weights, checkpoint hashes, epoch histories, and per-clip scores are retained. Input caching was checked for exact float32 equality without changing training or validation. All ten repartition seeds and their 1,000 fitted heads are retained.\par
The portable package verifies archived scores, refits primary frozen features, and supplies common-input/CNN/ViT primary-extraction tools. Authorized raw data, official weights, and pinned reviewed Transformer source are required separately. This is not a raw-data reproduction of all 148 historical conditions. Raw recordings and third-party weights/source are excluded from redistribution.\par

\end{document}